# Real-Time 2D Temperature Field Prediction in Metal Additive Manufacturing Using Physics-Informed Neural Networks

**Pouyan Sajadi** [1]
Email: sps11@sfu.ca

**Mostafa Rahmani Dehaghani** [1]
Email: mra91@sfu.ca

**Yifan Tang** [1]
Email: yta88@sfu.ca

**G. Gary Wang** [1]*
Email: gary_wang@sfu.ca

[1] Product Design and Optimization Laboratory, Simon Fraser University, Surrey, BC, Canada

## Abstract

Accurately predicting the temperature field in metal additive manufacturing (AM) processes is critical to preventing overheating, adjusting process parameters, and ensuring process stability. While physics-based computational models offer precision, they are often time-consuming and unsuitable for real-time predictions and online control in iterative design scenarios. Conversely, machine learning models rely heavily on high-quality datasets, which can be costly and challenging to obtain within the metal AM domain. Our work addresses this by introducing a physics-informed neural network framework specifically designed for temperature field prediction in metal AM. This framework incorporates a physics-informed input, physics-informed loss function, and a Convolutional Long Short-Term Memory (ConvLSTM) architecture. Utilizing real-time temperature data from the process, our model predicts 2D temperature fields for future timestamps across diverse geometries, deposition patterns, and process parameters. We validate the proposed framework in two scenarios: full-field temperature prediction for a thin wall and 2D temperature field prediction for cylinder and cubic parts, demonstrating errors below 3% and 1%, respectively. Our proposed framework exhibits the flexibility to be applied across diverse scenarios with varying process parameters, geometries, and deposition patterns.

---

∗ Corresponding Author.

field prediction, real-time prediction, physics-informed input, physics-informed loss function

## 1. Introduction

Metal additive manufacturing (AM) has emerged as a transformative technology, finding applications across a spectrum of industries, including aerospace, defense, and biomedicine. Its attractiveness stems from its capacity to create custom-designed 3D objects layer-by-layer, enabling mass customization, lightweight designs, efficient material recyclability, and shortened lead times (Shamsaei *et al.*, 2015).

In the metal AM process, rapid heating and cooling cycles typically occur, inducing substantial fluctuations in the temperature field within both the substrate and deposited layers (Thompson *et al.*, 2015). This dynamic thermal environment is a critical determinant of the final product's quality, leading to variations in microstructure (Bontha *et al.,* 2006; Lippold *et al.,* 2011), porosity (Barua *et al*., 2014), as well as stress and strain states (Shamsaei *et al*., 2015; H. Yan *et al*., 2018). Consequently, a precise understanding and comprehensive analysis of thermal conditions are imperative for effectively controlling the AM deposition process (Z. Yan *et al.,* 2018).

In this dynamic context, the integration of digital twins (DT) assumes a pivotal role, serving as a bridge between the physical and digital realms. A digital twin represents a virtual replica of the physical AM system, functioning in real-time synchronization with the actual manufacturing process. Within this framework, real-time or near-real-time prediction of temperature fields emerges not only as a pivotal element but also as a pathway towards optimizing parameters (Hosseini *et al.,* 2023), minimizing defects (Khairallah *et al.*, 2016; Ren *et al.*, 2019), ensuring consistent product quality (Zheng *et al*., 2008), reducing waste, enabling predictive maintenance, and continually enhancing the efficiency and reliability of metal AM technology (Zhang *et al*., 2020).

Physics-based computational models, such as the finite element method (FEM) and computational fluid dynamics (CFD), are extensively studied for predicting thermal behavior in AM, most of which are inherently multiscale and multiphysical (Razavykia *et al*., 2020). These models use partial differential equations (PDEs) like Navier–Stokes and heat transfer equations to understand thermal history and temperature distribution. FEM models, preferred by Roberts *et al.* (2009), offer computational efficiency over CFD, particularly for analyzing solid heat transfer (Denlinger and Michaleris, 2016; Liao *et al*., 2022a). Li *et al.* (2020) applied a thermal-fluid model integrating the level set method and Lagrangian particle tracking for laser powder-bed fusion (LPBF). Irwin *et al.* (2016) predicted the thermal history in directed energy deposition (DED) processes. Yan *et al.* (2018) crafted a thermal flow model for simulating flow dynamics in laser spot melt pools. Heigel *et al.* (2015) introduced a thermo-mechanical model for DED, incorporating forced convection to enhance precision. In another study, Bai *et al.* (2013) used temperature data from an IR camera to calibrate uncertain input parameters in thermal simulations, ensuring more accurate simulation input determination for weld-based additive manufacturing. Despite advancements in numerical models for simulating temperature distributions in metal AM processes, there exist several limitations that hinder the practical use in real-world applications, including real-time prediction, the necessity for deep physical and mathematical understanding, and the need for high-performance hardware (X. Ren *et al*., 2020).

As an alternative to physics-based numerical models, data-driven methods are employed in modeling the intricate behaviors of AM processes. The study of data-driven thermal prediction can be classified into two main categories based on their input variables. The first category comprises models that solely map process parameters and the properties of parts to temperature profiles. In contrast, the second category includes models that incorporate temperature data from adjacent elements or previous time steps during the AM process, accounting for thermal transfer effects. In the first category of studies, Roy and Wodo (2020) and Mozaffar *et al.*(2018) used FEM calibration and recurrent neural network (RNN)-based models to predict the local thermal history. These models can effectively forecast temperature fields inside fabricated parts and on their surfaces. In the second category, studies like (Paul *et al*., 2019; K. Ren *et al*., 2020; Stathatos and Vosniakos, 2019) utilize artificial neural networks and iterative prediction methods. They considered past temperatures, surrounding elements, and element locations relative to laser inputs, thus providing more accurate predictions for future time steps and various laser paths. Additionally, Tang *et al.* (2023) utilize temperatures from specific points on the printed layer to predict the complete temperature field for the yet-to-print layer. An artificial neural network predicts temperature profiles for these specific points, followed by the use of a reduced order model to reconstruct the temperature profile for the entire layer.

Data-driven models offer computational efficiency and reduced dependence on comprehensive physical knowledge. However, they exhibit certain drawbacks. They often operate as "black box" systems, lacking transparency in their decision-making processes. They require substantial amounts of data for effective training, which can be time-consuming and expensive, especially in the context of metal AM, where obtaining extensive and high-quality training data is costly or challenging.

Physics-Informed Neural Networks (PINNs) (Raissi *et al*., 2019) represent an innovative machine learning (ML) paradigm that integrates the laws of physics, typically described by PDEs, directly into the neural network architecture. PINNs have gained significant traction in recent years across various scientific domains, such as fluid mechanics (Jin *et al*., 2021), electromagnetic analysis (Noakoasteen *et al*., 2020), and crack recognition (Shukla *et al*., 2020), where they leverage physical principles to enhance predictive accuracy even when confronted with limited data. In the realm of metal AM, PINNs can accurately predict temperature fields and melt pool dimensions while optimizing computational efficiency (Hosseini *et al*., 2023; Li *et al*., 2023). As a pioneering work, Zhu *et al.* (2021) employed a PINN to predict temperature and melt pool fluid dynamics in LPBF by incorporating essential conservation equations and utilizing the finite element simulation data. These simulations run up to 2.0 milli-seconds but only a limited portion from 1.2 to 1.5 milli-seconds are used as labeled training data. In another study focused on the melt pool, Jiang *et al.* (2023) used the heat transfer equation alongside a small amount of data obtained from simulations to predict the temperature field and melt pool dimensions. Xie *et al.* (2022) integrated heat transfer laws into their PINN framework to predict temperature fields in single-layer and multi-layer DED processes. Their model surpasses data-driven methods in both scenarios, achieving an accuracy of over 90% with just 4000 training data points, in contrast to tens of thousands training points employed by most data-driven models. Moreover, in another study (Liao *et al*., 2022b), heat transfer principles and partially observed temperature data from infrared cameras were employed to accurately predict the temperature history and identify hidden parameters, such as the laser absorptivity, $C_P$, and the material thermal conductivity, within the process. This research also emphasizes the advantageous role of transfer learning techniques in enhancing training efficiency and boosting prediction accuracy.

The detailed information of studies on PINNs to predict temperature fields is summarized in Table 1. While significant progress has been made in integrating physics knowledge with ML models, these advancements are often constrained to modeling only a small fraction of the process, typically less than a second. Furthermore, a key limitation is the lack of real-time data integration in their modeling approaches, a crucial element that enables control and provides valuable insight into the thermal dynamics of the metal AM process. Lastly, current modeling schemes tend to be tailored to single tracks and constant geometries, falling short when applied to processes with varying geometries and deposition patterns.

Table 1. A summary of PINN studies for temperature field prediction of metal AM processes (MAPE is mean absolute percentage error, MRE is mean relative error and RMSE is root mean square error)

| Study | Process | Physics incorporated | Inputs | Predicted output | Number of data points | Accuracy |
|---|---|---|---|---|---|---|
| **Hosseini *et al.*, 2023** | Single-track LPBF | Thermal energy conservation equation | x, y, z, t, process parameters, material properties | Temperature field | NA | MAPE = 4.5% |
| **Shukla *et al.*, 2020** | Single-track laser metal deposition | Heat transfer in deposition and cooling stage | x, y, z, t | Temperature field | NA | MRE = 0.81% for three sample points |
| **Li *et al.*, 2023** | Single-track selective laser beam melting | Conservation laws of momentum, mass (Navier-Stokes), and energy | x, y, z, t | Melt pool temperature and dynamics | NA | MRE = 5.9% for melt pool length |
| **Zhu *et al.*, 2021** | Single-track DED | Heat transfer | x, y, z, t | Melt pool size and temperature distribution | NA | MRE = 2% |
| **Jiang *et al.*, 2023** | Single-track DED | Heat transfer | x, y, z, t | Temperature field | 30,000 simulation data | MRE = 2.05% for average of three multi-layer cases |
| **Xie *et al.*, 2022** | Single-track DED | Heat transfer | x, y, z, t | Temperature field | 100,000 simulation data | RMSE (K) with clean data = 3.59 |

To address these existing gaps, this paper introduces a novel physics-informed Convolutional Long Short-Term Memory (PI-ConvLSTM) framework. This framework leverages real-time temperature field data collected during the manufacturing process to predict the 2-D temperature field of the part for future timestamps. Unlike prior approaches, the proposed framework is versatile and applicable to scenarios involving diverse geometries, deposition patterns, and process parameters, enabling a real-time prediction of the manufacturing process.

The rest of the paper is organized as follows: Section 2 introduces the basics of PINNs and the governing equations employed for modeling. This Section further delves into each component of the proposed PI-ConvLSTM framework. In Section 3, the paper details two distinct applications used to validate the framework with discussions of the results for each application. Section 4 is dedicated to presenting insights, outlining limitations, and exploring prospects. Finally, Section 5 encapsulates the paper with a conclusion and suggests potential avenues for future work.

## 2. Methods

### 2.1. General PINN model

Conventional neural networks primarily operate by mapping input data to desired outputs through training and rely heavily on the quantity and quality of labeled data for effective learning. Meanwhile, in practical scenarios, physical systems are often governed by underlying physical laws represented by PDEs.

PINNs seamlessly integrate the PDEs and information from measurements into the neural network's loss function through automatic differentiation. In its general form, PDEs can be expressed as:

$$u_t(x,t) + N(u:\lambda) = 0 \, , x \in \Omega, t \in [0,T] \tag{1}$$

where $u_t(x,t)$ serves a dual purpose, acting as both the actual PDE solution and the target neural network's output. The term $N(u:\lambda)$ embodies a nonlinear differential operator, with $\lambda$ encapsulating relevant parameters within the PDEs. The spatial domain is denoted by $\Omega$, while $[0,T]$ represents the temporal extent. The core of PINNs lies the concept of residual as a quantifiable measure of PDE compliance. Mathematically, the residual of a PDE is expressed as:

$$Res = \hat{u}(x,t) + N(u:\lambda) \tag{2}$$

This residual encapsulates the underlying physical law represented by the PDEs. As the predicted output $\hat{u}(x,t)$ of the neural network approaches the actual solution $u_t(x,t)$, the residual progressively diminishes. Therefore, by incorporating this residual into the model's loss function, it serves as a regularization factor, effectively constraining the neural network to adhere more closely to the governing physical principles encoded by the PDEs. Additionally, PINNs incorporate boundary and initial conditions, represented by Equations (3), to establish a well-posed system.

$$u_t(x,t) - B(x,t) = 0 \, , x \in \partial\Omega \tag{3}$$

$$u_t(x,0) - I(x) = 0 \, , x \in \Omega \tag{4}$$

Here, $B(x,t)$ represents boundary conditions, and $I(x)$ signifies initial conditions. Integrating these physical laws into the loss function is essential for enforcing system constraints. These physical laws, alongside the supervised loss of data measurements (Equations (5)-(8)), become integral components of the neural network's loss function, resulting in Equation (9):

$$L_{pde} = \frac{1}{N_p} \sum_{i=1}^{N_p} Res_{pde}^2 \tag{5}$$

$$L_{ic} = \frac{1}{N_i} \sum_{i=1}^{N_i} (\hat{u}_i - I_i(x))^2 \tag{6}$$

$$L_{bc} = \frac{1}{N_b} \sum_{i=1}^{N_b} (\hat{u}_i - B_i(x,t))^2 \tag{7}$$

$$L_{data} = \frac{1}{N_d} \sum_{i=1}^{N_d} (\hat{u}_i - u_i)^2 \tag{8}$$

$$L_{total} = w_p L_{pde} + w_i L_{ic} + w_b L_{bc} + w_d L_{data} \tag{9}$$

Here, $L$ represents the loss function associated with each of the components in the provided equations; $Res_{pde}$ is the residual of the PDE and $\hat{u}_i$ is the neural network's output and $u_i$ represents the ground truth data points from the supervised learning dataset. In Equation 9, components are weighted using $w_p$, $w_i$, $w_b$, and $w_d$, respectively, and involve $N_p$, $N_i$, $N_b$, and $N_d$ sampling points associated with each respective loss term. The network is effectively trained by minimizing the loss, employing gradient-based optimizers like Adam (Kingma and Ba, 2014) and L-BFGS (Byrd *et al.*, 1995), all facilitated through the process of backpropagation. An overview of the PINN algorithm is depicted in Figure 1.

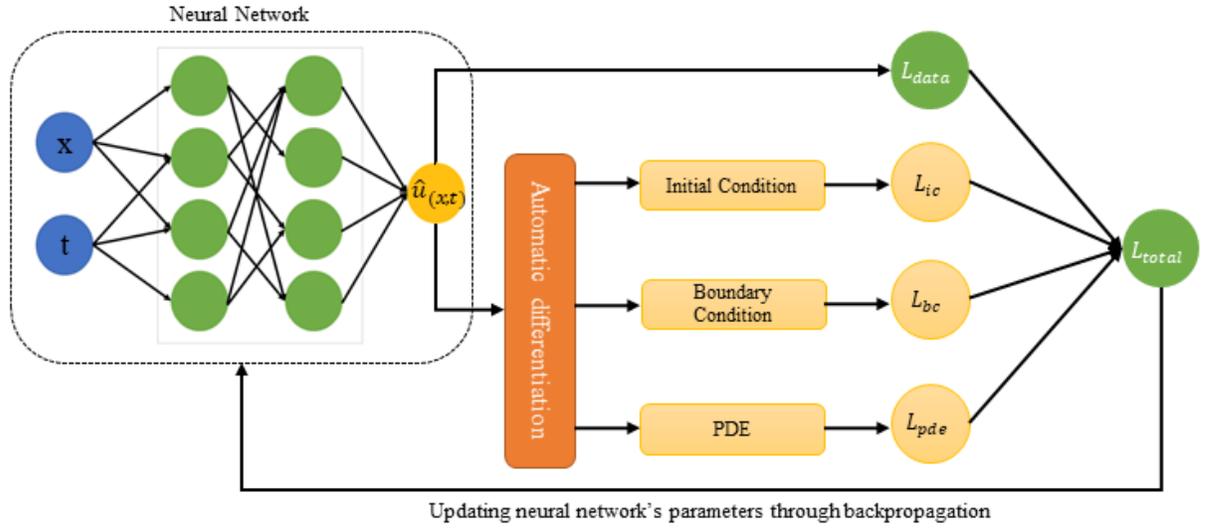

Figure 1. During training, the neural network predicts the output $\hat{u}(x,t)$, and computes the physics loss function, evaluating the system's adherence to physical laws. This loss function encompasses various components, including data-driven and physics losses, which are then minimized through backpropagation.

### 2.2. Governing Equations

In this Section, we introduce the governing equations for thermal prediction in this paper. Our focus is on modeling heat conduction in the metal AM process. With an exclusion of factors such as fluid flow and vaporization heat loss, the transient heat transfer during the process is represented by the following heat conduction PDE (Holman, 1986):

$$\rho C_p \frac{\partial (T(x,y,t))}{\partial t} = \frac{\partial}{\partial x}\left(k \frac{\partial T(x,y,t)}{\partial x}\right) + \frac{\partial}{\partial y}\left(k \frac{\partial T(x,y,t)}{\partial y}\right), \tag{10}$$

$$x, y \in \Omega, t \in [0, t_{end}]$$

where $T(x,y,t)$ is the corresponding temperature of position $(x,y)$ at time $t$, $\rho$ is the density of the part material, $C_p$ is the specific heat, and $k$ is the thermal conductivity. The initial condition is set equal to the ambient air temperature, $T_{amb}$. The boundary conditions are applied to all surfaces, except for the top surface where the laser heat flux is applied, accounting for the heat radiation and convection with the surrounding air.

For surfaces other than the top surface:

$$-k\frac{\partial T}{\partial \vec{n}} = h(T - T_{amb}) + \sigma\varepsilon(T^4 - T_{amb}^4), \qquad x \in \partial\Omega \tag{11}$$

For the top surface, with an additional term corresponding to the laser heat flux:

$$-k\frac{\partial T}{\partial \vec{n}} = h_c(T - T_{amb}) + \sigma\varepsilon(T^4 - T_{amb}^4) + Q_{laser}, \qquad x \in \partial\Omega_{top} \tag{12}$$

Here, $\frac{\partial T}{\partial \vec{n}}$ denotes the normal derivative perpendicular to the boundary. The coefficient $h$ signifies the heat convection coefficient characterizing the interaction between the substrate and air, while $\sigma$ stands for the Stefan–Boltzmann constant, and $\varepsilon$ represents the heat radiation coefficient. $Q_{laser}$ denotes the energy produced by the laser heat source per unit volume. These boundary conditions are essential for modeling the heat transfer and energy flow in the system.

### 2.3. Proposed Online PI-ConvLSTM Framework

In this section, we will explore the physics-informed ConvLSTM framework designed to address the challenging task of predicting the temperature field during the metal AM process. This framework leverages sequential thermal images and process parameters as input data to predict the 2D thermal field in subsequent timeframes. More specifically, a sequence of $w$ thermal images from timestamps $(t - w)$ to $t$ is input to the framework to predict the thermal image for the timestamp $(t + i)$. Here, $w$ represents the window size for the input, and $i$ denotes the timestamp in the future for which we aim to predict the temperature field. It's essential to note that both $w$ and $i$ are hyperparameters, and their significance will be elaborated on in the following sections.

This problem is inherently high-dimensional, given that the inputs and outputs consist of 2D thermal images. Furthermore, considering the temporal relationship between inputs and outputs due to the sequential nature of the data, we can view this problem as a 2D sequential modeling challenge. In

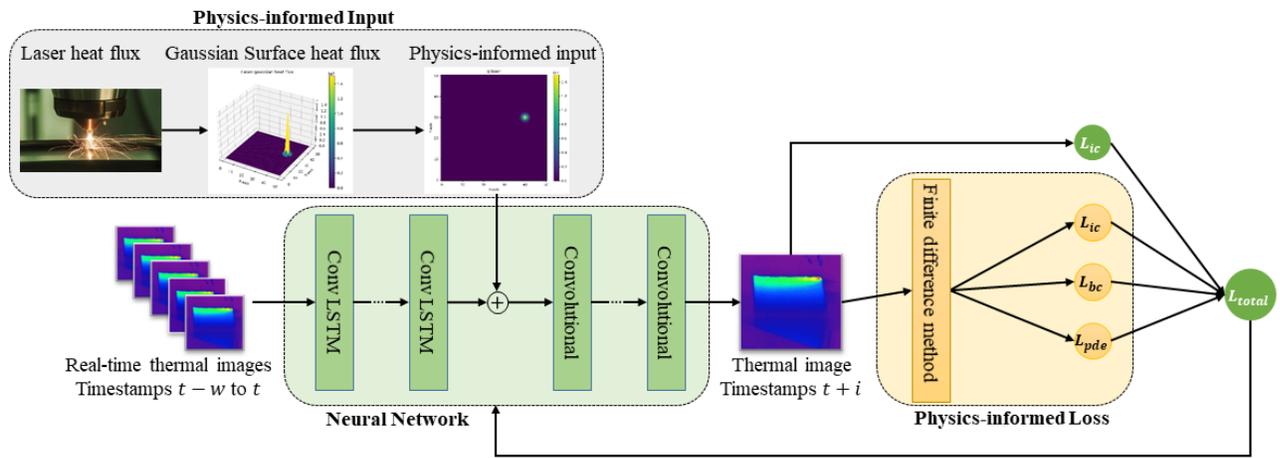

Figure 2, we provide an illustrative overview of our comprehensive framework, comprised of three key components: the neural network, Physics-informed (PI) loss, and Physics-informed (PI) input. These elements together underpin our strategy for addressing the complex challenge of predicting temperature distributions in a system with spatial and temporal dependencies. In the following sections, we will delve into each of these components, offering a closer look at their specific functions and how they work together within the framework.

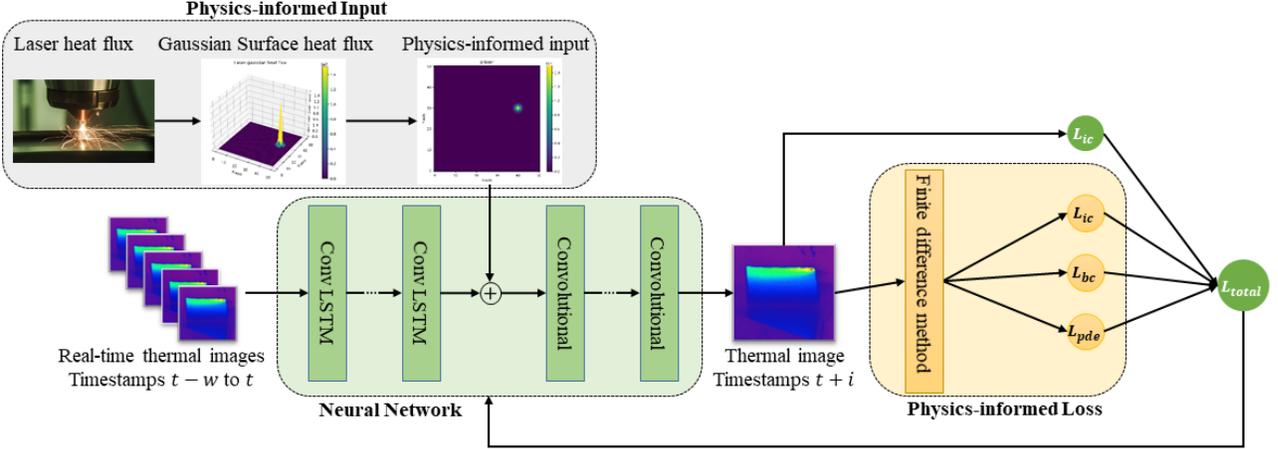

*Figure 2. Overview of PI-ConvLSTM framework with its components: the neural network, PI-input, and PI-loss*

### 2.3.1. Model Architecture

To tackle such complex and high-dimensional tasks, the choice of neural network architecture is crucial. In the literature, Convolutional Neural networks (CNN) are recognized for their ability to manage high-dimensional data by efficiently sharing parameters across layers. Spatial dependencies within 2D-to-2D modeling tasks are effectively captured by CNNs (Neupane *et al.*, 2021; Wu *et al.*, 2023; Zhou *et al.*, 2021), which employ convolutional layers to hierarchically extract features from the input data, enabling the learning of complex patterns.

Complementarily, the long short-term memory (LSTM) architecture excels at capturing temporal dependencies within sequential data (Sherstinsky, 2020) The recurrent nature of LSTM allows the maintenance of hidden states over time, rendering them suitable for modeling sequences and time series data (Yu *et al.*, 2019). LSTMs can capture patterns, trends, and dependencies in the temporal aspect of the data, which are critical information for predicting the temperature distribution across timestamps.

Building upon the strengths of CNNs and LSTMs, Convolutional LSTMs (ConvLSTM) have a hybrid architecture that seamlessly combines the spatial feature extraction capabilities of CNNs with the LSTM-like memory and sequential modeling capabilities (Shi *et al*., 2015). ConvLSTM operates by replacing the standard matrix multiplication in LSTM cells with convolutional operations. This fusion enables ConvLSTMs to simultaneously process spatial and temporal information, making them an ideal choice for tasks addressed in this paper, where 2D thermal images serve as inputs, and the goal is to predict a 2D temperature distribution evolving over time. The adoption of the ConvLSTM architecture equips the neural network with the capability to excel in the extraction of spatial features from thermal images through CNN-like operations, all while maintaining the capacity to model temporal dependencies using LSTM-like memory. Generally, the numbers of ConvLSMT modules and CNN modules depend on the complexity of tasks.

### 2.3.2. Physics-informed Loss Function

Referring to Section 2.1, a physics-informed loss function typically comprises two core elements: data-based loss and physics-based loss. The physics-based loss encompasses the residual of the PDE, quantifying the deviation between predictions and the PDE, along with the residual related to the boundary and initial conditions corresponding to the PDE.

The objective is to ensure that the framework's output adheres to the physics principles articulated by the PDE, as represented in Equations (10)-(12). To calculate temperature gradients across both time and space, the temperature distribution must be discretized into finite locations. This is achieved by subdividing the computation domain $\Omega$ into $m \times n$ cells with steps denoted as $h$. This discretization process is considered crucial when dealing with pixelized data, such as thermal images. Drawing inspiration from the finite difference method, we utilize differential approximations to compute the derivative terms in Equation 10.

$$\frac{\partial^2 T}{\partial x^2} \cong \frac{1}{h^2}[T(x_i + h, y_i) - 2T(x_i, y_i) + T(x_i - h, y_i)] \tag{13}$$

$$\frac{\partial^2 T}{\partial y^2} \cong \frac{1}{h^2}[T(x_i, y_i + h) - 2T(x_i, y_i) + T(x_i, y_i - h)] \tag{14}$$

The PDE loss can be formulated using the difference equations. Based on Equations (13) and (14), we derive the difference equation for Equation (10) as follows:

$$\begin{aligned} R(x_i, y_j) = &-\rho C_p \frac{\left(T(x_i, y_j) - T_{prev}(x_i, y_j)\right)}{\Delta t} \\ &+ \frac{1}{h^2} k \left(T(x_i + h, y_j) + T(x_i - h, y_j) + T(x_i, y_j + h) \right. \\ &\left. + T(x_i, y_j - h) - 4T(x_i, y_j)\right), \quad x \in \Omega, t \in [0, t_{end}] \end{aligned} \tag{15}$$

where $T(x_i, y_i)$ is the temperature of the pixel at $(x_i, y_i)$ in the 2D predicted temperature field and $T_{prev}(x_i, y_j)$ is the temperature of the pixel at $(x_i, y_i)$ for the previous timestamp. As the framework's outputs are high-dimensional images, the residual for each output is structured as follows:

$$Res_{pde} = \begin{bmatrix} R(x_0, y_0) & \cdots & R(x_n, y_0) \\ \vdots & \ddots & \vdots \\ R(x_0, y_m) & \cdots & R(x_n, y_m) \end{bmatrix} \tag{16}$$

In an ideal situation, the framework's output temperatures should drive $Res_{pde}$ to approach zero. Therefore, the PDE loss for the PI-ConvLSTM is formulated as:

$$L_{pde} = \frac{1}{N_p} \sum_{i=1}^{N_p} Res(i)_{pde}^2 \tag{17}$$

where $N_p$ is the number of thermal images used to train the framework. The losses associated with initial and boundary conditions remain consistent with the explanations provided in Section 2.1.

### 2.3.3. Physics-informed Input

In this section, the third component of the PI-ConvLSTM framework is introduced. Thus far, we have discussed the neural network architecture for predicting 2D temperature fields and explored the design of a specialized physics-based loss function. To improve the adaptability of our modeling approach for various manufacturing scenarios, we include important process parameters associated with the laser heat source.

Parameters such as laser power, laser absorptivity, beam radius, and laser location are deemed influential in thermal modeling for laser-based metal AM processes [15, 16]. Instead of directly providing these parameters as raw features to the model and expecting it to deduce their intricate relationship with temperature, we opt for a more informed strategy. We create an auxiliary physics-informed input, an intermediate feature that embeds meaningful physics knowledge related to the heat input, and incorporate it into a hidden layer of the neural network. This intermediate feature acts as a higher-level parameter that combines the influences of the mentioned process parameters. In our case, the selected higher-level parameter is the laser heat flux. The concept is to integrate information about the laser heat flux into the historical thermal information within the neural network to enhance the modeling process. In this study, we estimate the laser heat flux using a Gaussian surface heat flux model. The process of crafting this physics-informed input is detailed as follows: firstly, the heat flux for each point on the layer subjected to the laser's application is calculated as expressed in Equation (18):

$$q_{laser}(x,y) = -\frac{2\eta P}{\pi r_{beam}^2} \exp\left(\frac{2d^2}{r_{beam}^2}\right) \tag{18}$$

Here, $\eta$ represents the laser absorptivity, $P$ corresponds to the laser power, $r_{beam}$ stands for the laser beam radius, and $d$ represents the distance from the point $(x,y)$ to the laser center. Subsequently, we aggregate the input heat flux across the entire layer where the laser is applied, resulting in the formation of the heat flux matrix $q_{laser}$, as shown in Equation (19):

$$q_{laser} = \begin{bmatrix} q_{laser}(x_0, y_0) & \cdots & q_{laser}(x_n, y_0) \\ \vdots & \ddots & \vdots \\ q_{laser}(x_0, y_m) & \cdots & q_{laser}(x_n, y_m) \end{bmatrix} \tag{19}$$

Each element within the matrix $q_{laser}$ signifies the heat flux at a specific location on the layer where the laser is applied. Figure 3 illustrates an instance of laser application on a surface, modeled using the Gaussian surface heat flux, along with its corresponding $q_{laser}$ matrix. After this stage, the $q_{laser}$ matrix is introduced into the neural network following the ConvLSTM layers, where temporal dependencies in the data are extracted. Subsequently, within the neural network, multiple Convolutional layers are employed to extract spatial dependencies. This innovative approach allows us to effectively integrate crucial process parameters into our modeling framework, offering an enhanced understanding of the complex relationship between these parameters and temperature distributions.

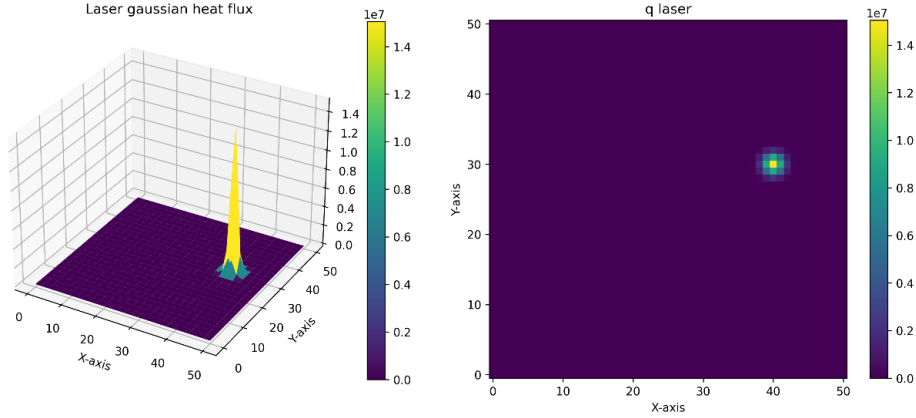

Figure 3. Gaussian heat flux on a surface (left) and its corresponding $q_{laser}$ matrix (right)

## 3. Applications

In this Section, we delve into the practical applications of the PI-ConvLSTM framework. To evaluate its efficacy and flexibility in thermal field prediction, we present two distinct scenarios. Firstly, we examine its capacity to forecast a 2D temperature field within a thin wall during a metal AM process, utilizing available experimental data from the literature. Secondly, we assess its performance in predicting 2D temperature fields for the actively printed layer, covering both cylindrical and cubical geometries, based on simulation data. This approach allows us to evaluate the framework's robustness and reliability in diverse data contexts.

### 3.1. 2D-Temperature Field Prediction of a Thin Wall

#### 3.1.1. Dataset and Experimental Setup

In the first application, we use the infra-red images taken during the LMD process for printing 60-layer Ti-6Al-4V thin-walled structures, provided by Marshall *et al.* (2016a). These data are measured by an IR camera, which is a part of the OPTOMEC Laser Engineered Net Shaping (LENS) 750 printer. The system setup is shown in Figure 4. The thin wall was constructed at an orientation such that one of its sides was fully in-view by the IR camera. The data of the IR camera are output to comma separated value (CSV) files, each of which contains a 320×240 (width × height) matrix of temperature values. In total, 2760 thermal images are captured by the IR camera during the deposition process. These images undergo processing and are organized so that a sequence of $w$ images forms one input, with the subsequent image corresponding to the next timestamp as the output. For instance, images from timestamp 100 to timestamp $(100 + w)$ serve as input to the framework, predicting the output image for timestamp $(100 + w + 1)$. As mentioned in Section 2.3, $w$ serves as the window size for inputs and is determined through hyperparameter tuning.

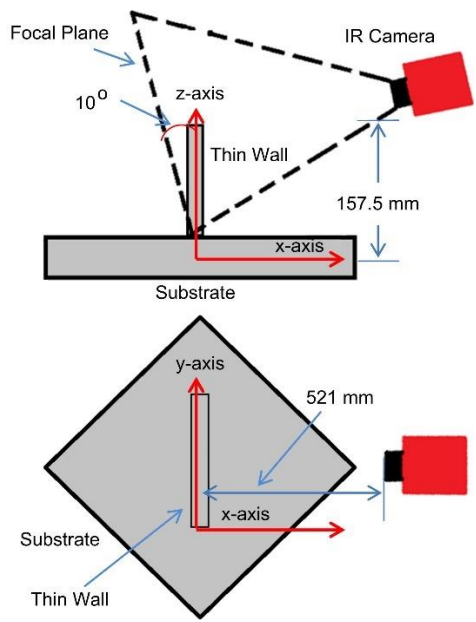

Figure 4. Side view (top) and aerial view (bottom) of IR camera and its orientation with respect to the substrate and thin wall within the build chamber (Marshall *et al*., 2016b).

### 3.1.2. PI-ConvLSTM Framework for Full-Field Temperature Prediction

The developed PI-ConvLSTM framework serves as a tool for predicting the full temperature field for the i[th] future timestamp based on thermal images captured at previous timestamps. In this specific application, we simplify the model by assuming that temperature remains uniform throughout the material's thickness, thus allowing us to disregard solid heat transfer in the through-the-thickness direction. This simplification transforms the heat transfer problem within the thin wall into a 2D case. In this revised model, we treat the convective and radiative heat flux occurring on the two surfaces of the wall parallel to the yz plane as a heat source term. This leads to the following expression of Equation 10:

$$\rho C_p \frac{\partial (T)}{\partial t} - k \left( \frac{\partial^2 T}{\partial x^2} + \frac{\partial^2 T}{\partial y^2} \right) + \frac{h}{w}(T - T_{amb}) + \frac{\sigma \varepsilon}{w}\left(T^4 - T_{amb}^4\right) = 0 \qquad (20)$$

Here, *w* represents the thickness of the wall. The boundary conditions specified in Equation (11) are still valid for the left and right boundaries of the 2D wall, and Equation (12) applies to the upper surface of the wall.

The input comprises a series of previous IR images and the framework is informed by heat transfer dynamics and associated boundary conditions, which enables it to predict the full-field temperature at the next timestamp. Additionally, laser heat flux data, as described in Section 2.3.3, is incorporated into the model to enhance its understanding of the relevant process parameters. In the model's architecture, three sequential ConvLSTM layers and two Convolutional layers, each equipped with 10 filters, are employed. The model is trained for 40 epochs using the Adam optimizer with a learning rate of 1e-3. All implementations are performed using Python's TensorFlow package.

### 3.1.3. Results and Discussion for Full Field Temperature Prediction

To assess the performance of the developed framework against experimental data, we utilize various evaluation metrics, including Mean Absolute Error (MAE), Mean Square Error (MSE), and Mean Absolute Percentage Error (MAPE). The use of these three metrics provides a comprehensive evaluation of the model's performance. MSE emphasizes overall accuracy with a focus on larger errors, while MAE treats all errors equally, and MAPE gauges relative accuracy by expressing errors as a percentage of actual values. This multi-metric approach ensures a nuanced understanding of precision, error magnitude, and percentage-wise accuracy, enhancing the overall assessment of the model.

$$MSE = \frac{1}{n \times m} \sum_{i=1}^{n} \sum_{j=1}^{m} (Y_{ij} - \hat{Y}_{ij})^2 \tag{21}$$

$$MAE = \frac{1}{n \times m} \sum_{i=1}^{n} \sum_{j=1}^{m} |Y_{ij} - \hat{Y}_{ij}| \tag{22}$$

$$MAPE = \frac{1}{n \times m} \sum_{i=1}^{n} \sum_{j=1}^{m} \left| \frac{Y_{ij} - \hat{Y}_{ij}}{Y_{ij}} \right| \tag{23}$$

where, $Y_{ij}$ and $\hat{Y}_{ij}$ represent the actual and predicted temperature fields for each element in the dataset respectively, with $n$ and $m$ denoting the number of rows and columns in the temperature matrix. These metrics assess individual errors. The overall error for validation dataset is obtained by averaging these individual errors, providing an evaluation of the entire validation dataset.

- **Effect of window size on prediction error**

In Section 2.3, the window size is identified as a hyperparameter to be determined through tuning. To find the most accurate window size, a range of window sizes from $w = 1$ to 6 is tested, and the outcomes are depicted in Figure 5. Based on these results, a window size of five demonstrates the lowest MSE, hence is chosen as the number of previous timestamps to use as input.

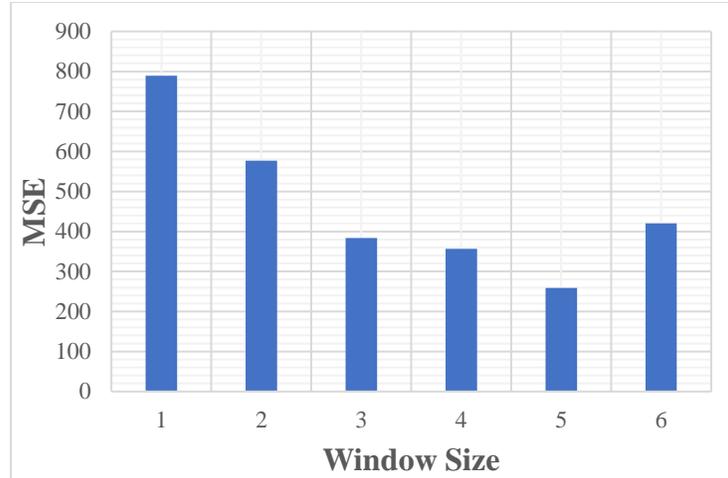

Figure 5. MSE on the validation dataset for different window sizes

- **2D temperature prediction results**

Figure 6 shows a comparison between the predicted temperature field made by the PI-ConvLSTM framework and the actual experiment results at different times. It also displays the difference between the predicted and experimental data. While most areas in the prediction closely match the real results, accuracy slightly decreases around the melt pool region. This reduced accuracy can be attributed to the intricate and complex nature of the melt pool area. The area's swift temperature fluctuations, complex phase transitions, and dynamic material behavior present challenges for precise prediction due to their highly nonlinear and transient characteristics. Moreover, the melt pool is subject to diverse boundary conditions, such as extreme temperatures and rapid thermal gradients, which, if not accurately represented, significantly impact the model's predictive accuracy within this region.

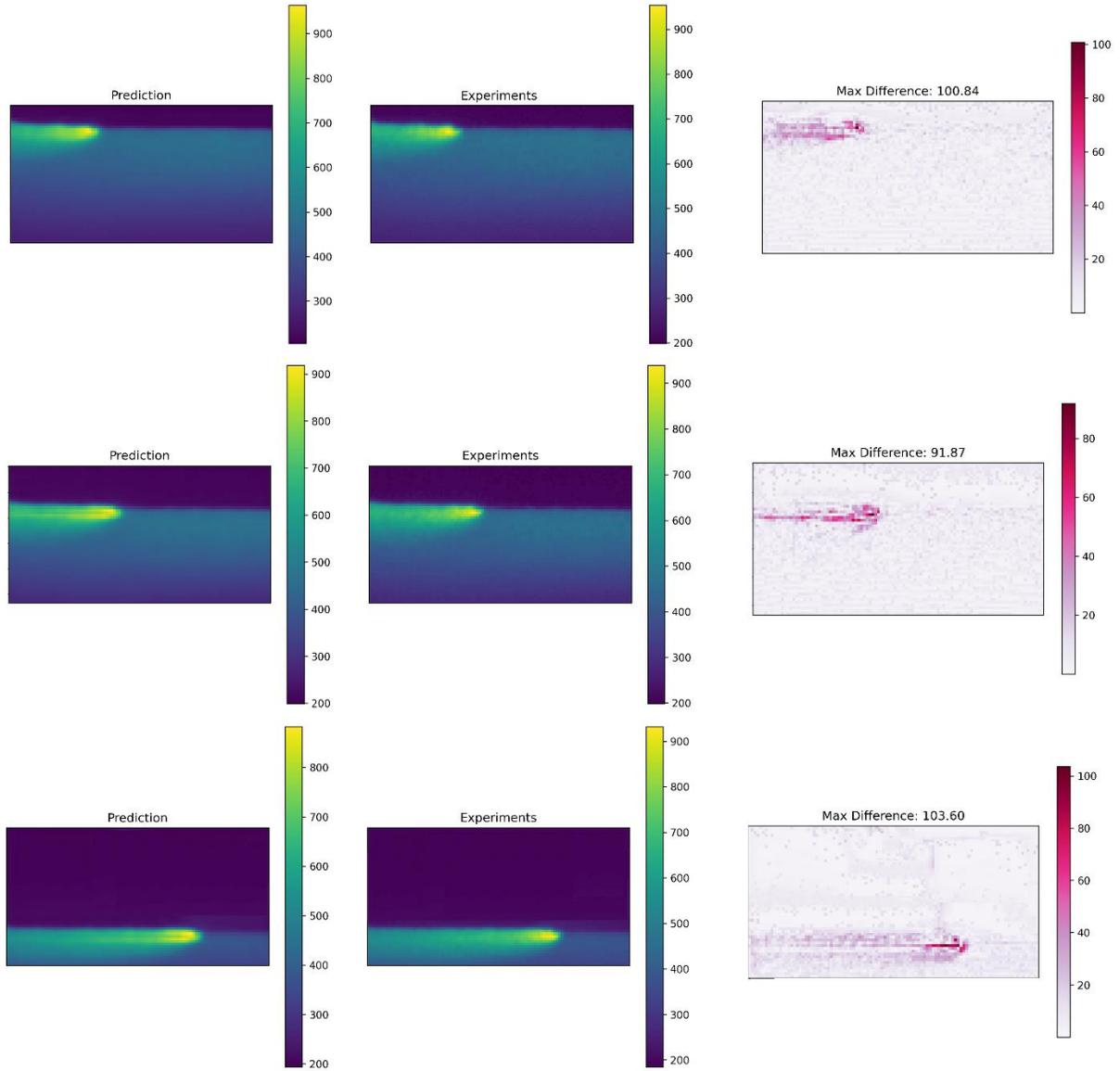

(a) Framework prediction     (b) Experimental result     (c) Absolute difference between framework prediction and experimental result

Figure 6. Comparison of the temperature field from the framework prediction and experimental results in a cross-Sectional view (top timestamp=2300 (t=391s), middle timestamp=1900 (t=270s), below timestamp=900 (t=117s))

- **Effect of Physics-informed components on prediction error**

As previously described, the PI-ConvLSTM framework consists of three integral components: the PI loss, the PI input, and a neural network. In order to evaluate the influence of each component, we conducted multiple experiments. We trained the model in various configurations: one included both the PI loss and PI input, another solely used the PI loss, a third model only implemented the PI input, and the final model (ML-only model) excluded any PI components. The obtained results, shown in Table 2, indicated that the model utilizing solely the PI input excelled over the ML-only model across various evaluation metrics. Additionally, the model incorporating the PI loss displayed a strong performance, benefiting from its ability to integrate the PDE, boundary condition, and initial condition to enhance robustness against existing noises within the data. Notably, the model that combined both the PI loss and PI input (i.e., PI-ConvLSTM) outperformed other configurations across all evaluation metrics. This superiority is due to its comprehensive integration of physics principles through the PI loss, inclusion of process parameters like laser heat flux via the PI input, and utilization of experimental information for comprehensive field temperature prediction.

Table 2. The results for PI-ConvLSTM frameowrk with different configurations

| Model Variation | MSE | MAE | MAPE |
| --- | --- | --- | --- |
| ML Only | 1967 | 57.5 | 14.39% |
| PI input | 1176 | 48.9 | 11.14% |
| PI loss | 302 | 13.1 | 2.73% |
| PI input + PI loss | 273 | 11.3 | 2.15% |

### 3.2. 2D-Temperature Field Prediction of a Cylinder and a Cube

#### 3.2.1. Dataset and Simulation Setup

In the second application, we employ the PI-ConvLSTM framework to predict 2D temperature field for the currently printed layer of a cylinder and a cubic shape part. Within this section, to test the adaptability of the framework, two distinct geometries with corresponding deposition patterns have been selected. Cylindrical and cubic geometries, widely used in the industry are chosen to serve as the foundation for the study. An overview of these geometries and deposition patterns can be observed in Figure 7.

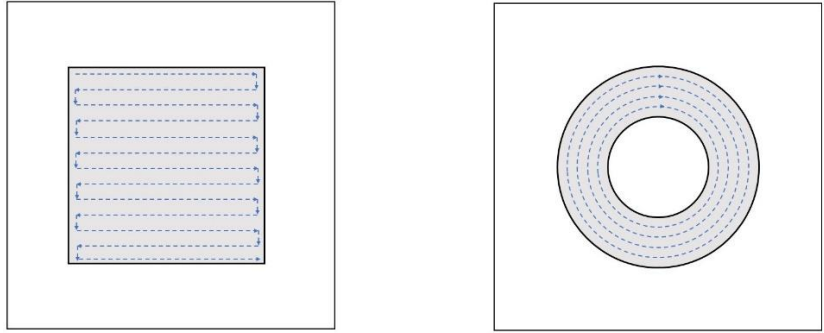

Figure 7. Illustrations of geometry and deposition patterns for 2D temperature field prediction. The grey lines indicate the base, the black lines indicate the deposition geometry, and the dashed arrows indicate the direction of the laser.

Table 3. Material properties used in the simulation

| Parameter | Value | Unit |
|---|---|---|
| **Density, $\rho$** | $7915 - 0.59 \times T$ | $kg/m^3$ |
| **Heat conductivity, $k$** | $12.6 + 0.015 \times T$ | $W/m\,C$ |
| **Heat capacity, $C_p$** | $496.5 + 0.133 \times T$ | $j/kg\,C$ |
| **Convection coefficient, $h$** | 10 | $W/m^2\,C$ |
| **Emissivity, $\varepsilon$** | 0.3 | - |

We utilized the pre-built formulations and integrated physics within the ANSYS AM DED process module, specifically crafted for the Directed Energy Deposition process for the simulation of the process. In the simulations, the pass width and the layer thickness are set to maintain a consistent pass of 1 mm. Both the substrate and deposited parts are made from the 316 stainless steel. A summary of material properties can be found in Table 3. Our thermal analysis encompasses considerations such as thermal conductivity, thermal convection, and radiation. The initial temperature of the substrate and ambient temperature is consistently set at 23°C. The deposition speed stands at 10 mm/s for the cylinder and 7 mm/s for the cubic part. The ANSYS AM DED simulation employs an inactive activation strategy, where elements representing the deposited material are activated following the deposition sequence at specific times. Upon activation, these elements are assigned a temperature defined as the "process temperature". The determination of the process temperature involves referencing empirical data from similar cases. In this scenario, the assigned values for element activation are 1800°C for the cubic part and 2400°C for the cylindrical part. Notably, the simulation does not model the laser heat flux. Furthermore, the laser power is treated as a trainable parameter and adjusted during the training process. Each simulation generates a dataset for every timestamp. In each dataset, the transient temperature values for all nodes are recorded. These datasets are then preprocessed to extract specific points of interest. Consequently, the input-output pairs for training and validation are structured as follows: a sequence of the 2D temperature field for the currently printed layer and the layer below are used as inputs, with the corresponding output representing the 2D temperature field of the currently printed layer for the next timestamp. Each simulation comprises roughly 16,000 timestamps, resulting in approximately 16,000 input-output pairs extracted for training and validating the framework for each of the geometries. 80% of the data is allocated for training, while the remaining 20% is set for validation.

### 3.2.2. PI-ConvLSTM Framework for 2D Temperature Field Prediction

In this application, our focus lies on predicting the 2D temperature field for the currently printed layer. To achieve this, the model exclusively integrates the boundary conditions related to the top surface, where the laser is applied. Therefore, Equation (12) is included in the PI loss function, which can be represented as:

$$L_{total} = w_b L_{bc} + w_i L_{ic} + w_d L_{data} \tag{24}$$

$w_b$, $w_i$ and $w_d$ are weights set proportionally to ensure a balanced scale among the various terms in the loss function, contributing to the model's enhanced training robustness. This tailored approach allows the model to work with image-based inputs for various geometries. Due to the complexity of the geometry and availability of more training data, a more sophisticated model is used for the second application. This approach employs a model equipped with six ConvLSTM layers and four Convolutional layers, each utilizing 20 filters, to predict the 2D temperature field for both cylindrical and cubic parts. The laser heat flux distribution for each geometry at every timestamp is integrated into the model after the ConvLSTM layers. Each geometry undergoes separate training sessions consisting of 12 epochs, utilizing the Adam optimizer with a learning rate of 1e-3.

### 3.2.3. Results and Discussion for 2D Temperature Field Prediction

- **Effect of the window size on the prediction error**

To determine the optimal window size for 2D temperature field prediction, multiple window sizes from $w = 1$ to 7 were tested. From the outcomes detailed in Figure 8, a window size of three was found to yield the lowest MSE, and subsequently selected as the number of preceding time steps for the model input.

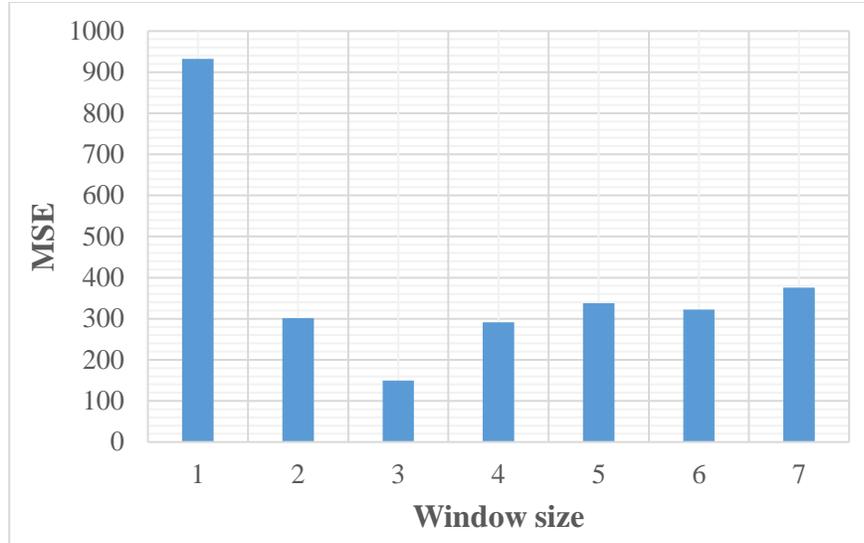

Figure 8. MSE on the validation dataset for different window sizes

- **2D temperature prediction results**

Consecutively, the results for 2D temperature field prediction for the cylinder and cubic parts are presented in Table 4. These findings reveal a strong concordance between the predictions generated by the PI-ConvLSTM framework and the simulation results. The predictions demonstrate less than 1% MAPE and an approximate MAE of 7°C for both geometries and deposition patterns. While the relative error in cylinder geometry is lower than in the cube, the increased temperature range in this process contributes to generally higher MSE and MAE for the cylinder, with the squared error notably intensifying in higher temperature ranges, particularly in the MSE. The accuracies in both cases are notably satisfactory, indicating the robustness and adaptability of the framework for temperature predictions across diverse process parameters, geometries, and deposition patterns.

Table 4. Prediction results of 2D temperature fields using the PI-ConvLSTM framework

| Geometry-deposition pattern | MSE | MAE | MAPE |
|---|---|---|---|
| Cylinder-Spiral | 149.5 | 7.62 | 0.67% |
| Cube-ZigZag | 115.0 | 6.85 | 0.74% |

Figure 9 compares the predicted temperatures from the PI-ConvLSTM with simulated temperatures for both the cylinder and cube. The predicted temperatures match well with the actual temperatures in most areas, with differences typically below 20°C. Closer to the melt pool, however, there are bigger differences, likely for reasons similar to those in Figure 6. Notably, the model's predictions are less accurate for the cylinder than for the cube, mainly because of the higher process temperature chosen for this geometry. Additionally, the more complicated pattern of material deposition in processing the cylinder may cause additional errors in modeling how the heat is transferred.

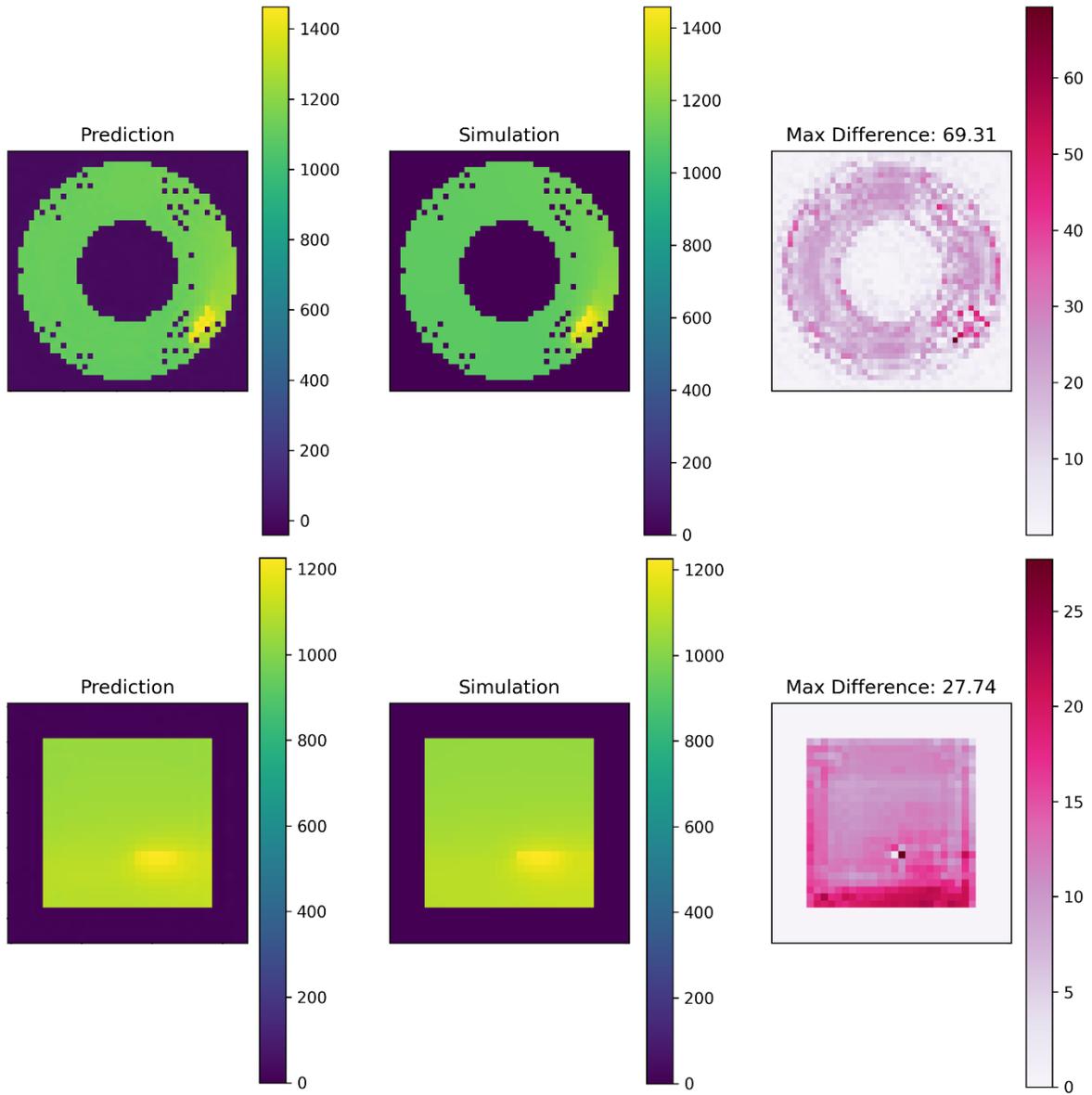

(a) Framework prediction  (b) Simulation result  (c) Absolute difference between framework prediction and simulation result

Figure 9. Comparison of the temperature field from the framework prediction and simulation results in a top view

(top – the cylindric part, bottom – the cubic part)

- **Evolution of loss terms through training**

To assess the framework's performance in data-based and physics-based aspects, Figure 10 illustrates the evolution of boundary condition and data loss across the training process. The decreasing data loss over training suggests the model is adapting to the data. Boundary condition loss not converging to zero, however, suggests that while the physics-based constraints in Equation 12 provide valuable insights into the thermal behavior in the process, they might not be comprehensive in describing the entire thermal system. The convergence to a non-zero value implies that merely enforcing physics principles might not be adequate on their own to achieve the desired solution. Therefore, the combination of data loss with the underlying physics principles becomes imperative for a more accurate representation of the system dynamics.

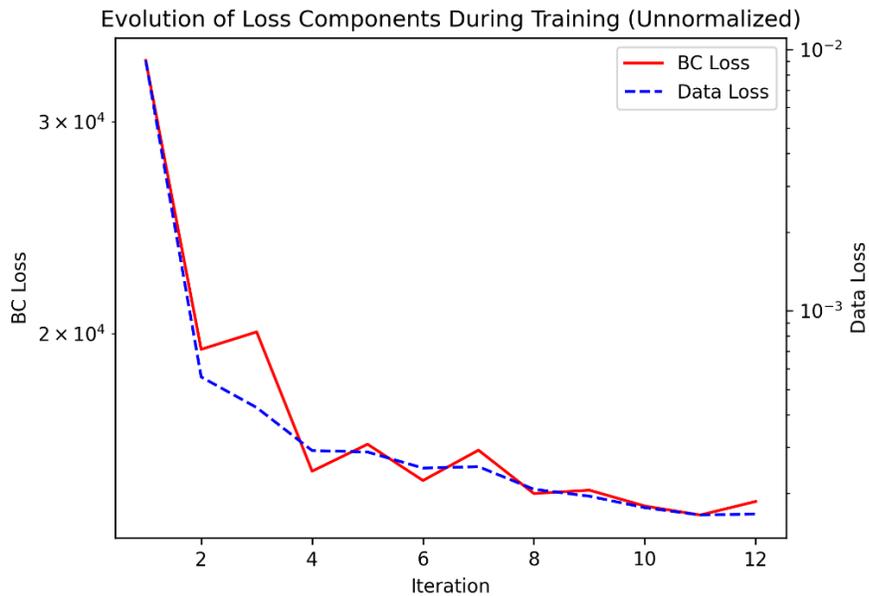

Figure 10. Evolution of unweighted loss terms during the training process

- **Residual of physics-based loss function through training**

Figure 11 demonstrates the variation in the residual matrix associated with the boundary condition loss as the training progresses, for a sample test input. Each matrix within the figure represents the residual values corresponding to a distinct iteration during the training of the model. These matrices depict the residuals across the surface of the dataset: brighter points denote higher residuals, while dark shades indicate lower residuals. Initially, in the early stages of training, the matrices predominantly exhibit higher residuals, signifying disparities between the model's predictions and the standards defined in the physics-based loss. As training progresses, the matrices gradually exhibit lower residuals, indicating improved alignment between the model's prediction and the governing physics principles (as represented by Equation 12).

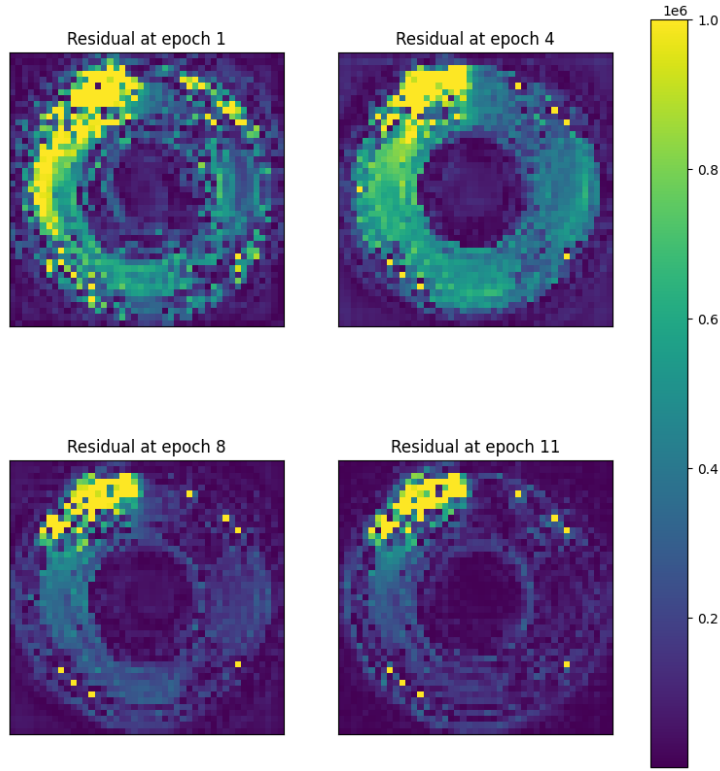

Figure 11. Residual of the Boundary Condition loss during the training process

- **Effect of the number of training data**

To explore the influence of training data quantity on the framework's accuracy, the model was trained using 800, 3200, 5600, 8000, and 12500 pieces of training data. As illustrated in Figure 12, a noteworthy finding emerged: achieving below 10% error required just 3,200 training samples, while obtaining less than 1% error for predicting the next timestamp required 12,500 training samples. This result emphasizes the framework's efficiency in delivering accurate predictions with a limited dataset. The result highlights that the framework efficiently provides accurate predictions despite using a limited dataset, showcasing its robustness and effectiveness in practical scenarios when available data is limited.

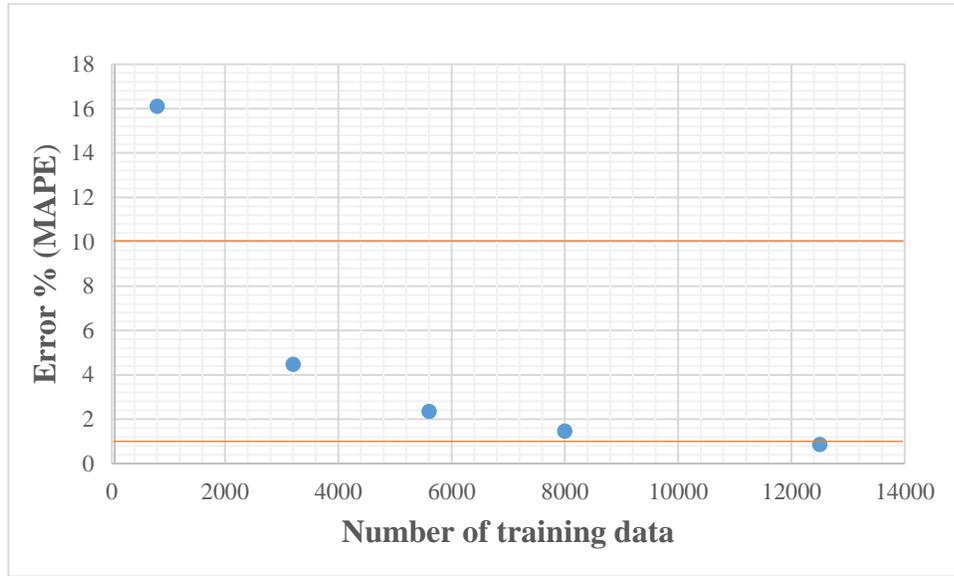

Figure 12. PI-ConvLSTM prediction error with different numbers of training data

- **Prediction of extended timestamps**

In practical applications, the utility of the framework relies on its ability to not only predict the temperature field for the next timestamp but also for more extended future periods. To assess the accuracy of the framework in predicting later timestamps, two distinct approaches are considered. The first approach involves a rolling prediction, where the prediction for the timestamp $(t + 1)$ serves as input for predicting $(t + 2)$, and this process continues iteratively. In this approach, the hyperparameter $i$, representing the timestamp in the future for prediction, is set to 1, and the framework iteratively predicts the next timestamp. The second approach is a direct prediction of a future timestamp (e.g., $t + 10$) by setting $i$ to the desired prediction timestamp. For example, with $i = 10$, the framework predicts the thermal image (i.e., the 2D temperature field) for the 10th timestamp in the future.

The results for both approaches are depicted in Figure 13 (a) and Figure 13 (b). In the rolling approach, prediction errors remain below 8% for timestamps up to the 10th, but its accuracy drastically decreases for timestamps beyond the 10th. As for the direct approach, prediction errors for $i = 10, 20, 50, 100$, and 200 are illustrated. The prediction error increases as we attempt to predict timestamps further in the future, although it remains lower than that using the rolling approach. This difference arises because, in the rolling approach, where predictions for later timestamps depend on preceding ones, the error accumulates over time, limiting its effectiveness for predicting many timestamps in the future. Conversely, the direct approach avoids error accumulation, making it more useful for predicting timestamps far in the future.

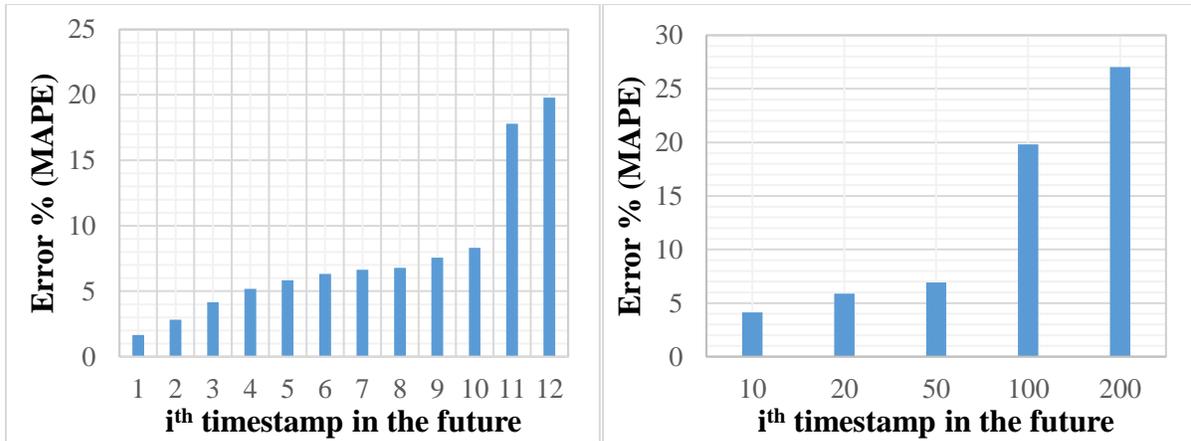

Figure 13. a) Error for rolling prediction of the $i^{th}$ timestamp in the future

b) Error for direct prediction of the $i^{th}$ timestamp in the future

However, if the focus is on understanding the dynamics of thermal behavior, the rolling approach can provide insights at a lower computational cost, as the direct approach necessitates training separate models for each timestamp $i$. In conclusion, for predicting the temperature field for longer timestamps in the future, the direct approach demonstrates a higher potential. However, if the objective is to comprehend the temporal dynamics of the temperature field, the rolling approach can yield results by training only a single model for the next timestamp (i.e., $i = 1$).

## 4. Insights, Constraints, and Prospects

The proposed framework demonstrates a notable capacity to predict the thermal field with a precision that yields an error rate below 3% for the subsequent timestamp and below 7% when forecasting up to the 50th timestamp for the cylinder part. This capability suggests its potential integration into control systems to facilitate corrective actions by adjusting process parameters to ensure process stability. The architecture of the ConvLSTM, along with the design of the loss function, offers flexibility in predicting temperature fields for different geometries and diverse deposition patterns through a single unified model, suggesting its potential in broader applications. Notably, its utility extends to systems requiring future timestamp predictions. This model can potentially substitute field-governing PDEs with the heat transfer PDE and laser heat flux employed in this study, allowing for wider applicability across various domains.

Despite these advancements, the framework faces inherent limitations. While it incorporates the heat transfer equation in the loss function, the physics-informed loss converges to a non-zero value, indicating the PDEs are not sufficient in describing the thermal behavior within the process. Furthermore, the resolution of the discretization, both spatially and temporally, significantly influences the accuracy of predictions. Moreover, the approach relies on real-time temperature data acquisition through an infrared (IR) camera, which could be a practical constraint, limiting its applicability in processes without such sensors.

## 5. Conclusions

This study introduces the PI-ConvLSTM framework, a physics-informed neural network designed for real-time temperature field prediction in metal additive manufacturing (AM) processes. The framework employs transient heat transfer equations and a Gaussian surface heat flux model as physics-informed loss and input, respectively, for a neural network with convolutional and convolutional LSTM layers. The proposed approach predicts the temperature field for future timestamps using real-time temperature field data from previous timestamps. Validated with three geometries—a thin-walled structure, a cylindrical part, and a cubic part—the highlights of this work are summarized as follows:

- Integrating residuals of transient heat transfer equations, boundary condition, and initial condition as a physics-based regularization penalty improves model accuracy compared to the loss solely based on data.
- Instead of inputting raw process parameters directly into the model, a Physics-informed input—laser heat flux in this paper—is utilized to better capture the relationship between process parameters and the temperature field. This Physics-informed input demonstrates a positive impact on the model's accuracy.
- The proposed framework achieves high accuracy, with an error below 3% for full field temperature prediction in a thin-walled part and below 1% for 2D temperature fields in cylindrical and cubic parts for subsequent timestamps. The tests demonstrate the flexibility of the proposed framework in handling diverse scenarios with varying process parameters, geometries, and deposition patterns.

- The proposed framework excels in predicting temperature fields for extended future periods, employing both rolling and direct prediction approaches. The direct approach demonstrates superior potential for longer-term predictions, while the rolling approach provides insights into thermal dynamics at a lower computational cost.

Future work will focus on advancing the current framework to enable online training, allowing for the continual updating of model parameters using real-time data acquired during the manufacturing process.

## CRediT authorship contribution statement

**Pouyan Sajadi:** Conceptualization, Investigation, Writing – original draft, Methodology, Visualization, Writing – review & editing. **Yifan Tang:** Conceptualization, Writing – review & editing. **Mostafa Rahmani Dehaghani:** Conceptualization, Writing – review & editing. **G. Gary Wang:** Supervision, Funding acquisition, Conceptualization, Writing – review & editing,

## Declaration of Competing Interest

The authors declare that they have no known competing financial interests or personal relationships that could have appeared to influence the work reported in this paper.

## Declaration of Generative AI and AI-assisted technologies in the writing process

In the process of preparing this work, the authors utilized ChatGPT to make subtle refinements, contributing to the overall polish of the text. After incorporating this tool, the authors conducted a careful review and made necessary edits to ensure the publication's quality, taking full responsibility for the final content.

# Acknowledgements

Funding from the Natural Science and Engineering Research Council (NSERC) of Canada under the project RGPIN-2019-06601.